%% file: arxiv.tex
\documentclass[twoside]{article}

\usepackage[round]{natbib}

\usepackage[OT1]{fontenc} % Use 8-bit encoding that has 256 glyphs
\usepackage[utf8]{inputenc}
\usepackage{graphicx}
\usepackage[export]{adjustbox}
\usepackage{booktabs} % Horizontal rules in tables
\usepackage{tabularx}
\usepackage{hyperref} % For hyperlinks in the PDF
\usepackage{amsmath,amsthm,amssymb,amsfonts}
\usepackage{microtype}
\usepackage{cleveref}
\usepackage{subcaption}
\usepackage{placeins}
\usepackage{multirow}

\usepackage[left=0.75in,right=0.75in,top=1in,bottom=1in]{geometry}

\usepackage{libertine}

\crefname{equation}{equation}{equations}
\Crefname{equation}{Equation}{Equations}% For beginning \Cref
\crefrangelabelformat{equation}{(#3#1#4--#5#2#6)}

\crefmultiformat{equation}{(#2#1#3}{, #2#1#3)}{#2#1#3}{#2#1#3}
\Crefmultiformat{equation}{(#2#1#3}{, #2#1#3)}{#2#1#3}{#2#1#3}

\newcommand{\x}{\mathbf{x}}

\newcommand{\A}{\mathbf{A}}
\newcommand{\K}{\mathbf{K}}

\newcommand{\y}{\mathbf{y}}
\newcommand{\f}{\mathbf{f}}
\newcommand{\w}{\mathbf{w}}
\newcommand{\g}{\mathbf{g}}
\newcommand{\m}{\mathbf{m}}

\newcommand{\F}{\mathbf{F}}
\newcommand{\U}{\mathbf{U}}
\newcommand{\Z}{\mathbf{Z}}
\renewcommand{\l}{\ell}

\newcommand{\0}{\mathbf{0}}

\newcommand{\bmu}{{\boldsymbol\mu}}

\newcommand{\E}{\mathbb{E}}
\newcommand{\R}{\mathbb{R}}

\newcommand{\N}{\mathcal{N}}
\newcommand{\X}{\mathbf{X}}

\newcommand{\GP}{\mathcal{GP}}

\renewcommand{\u}{\mathbf{u}}

\renewcommand{\S}{\mathbf{S}}
\newcommand{\z}{\mathbf{z}}

\def\bk{\mathbf{k}}

\DeclareMathOperator{\KL}{KL}

\DeclareMathOperator{\cov}{\textbf{cov}}

\title{Deep convolutional Gaussian processes}
\author{
    Kenneth Blomqvist \qquad Samuel Kaski \qquad Markus Heinonen \\ 
    Department of Computer Science, Aalto university \\ 
    Helsinki Institute for Information Technology HIIT
}

\begin{document}
\twocolumn[\maketitle]

\begin{abstract}
  We propose deep convolutional Gaussian processes, a deep Gaussian process architecture with convolutional structure. The model is a principled Bayesian framework for detecting hierarchical combinations of local features for image classification. We demonstrate greatly improved image classification performance compared to current Gaussian process approaches on the MNIST and CIFAR-10 datasets. In particular, we improve CIFAR-10 accuracy by over 10 percentage points.
\end{abstract}

\section{Introduction}
\label{sec:introduction}
\input{01-introduction.tex}

\section{Background}
\label{sec:background}
\input{02-background.tex}

\input{03-methods.tex}

\section{Experiments}
\label{sec:experiments}

\input{04-experiments.tex}

\section{Conclusions}
\label{sec:conclusions}
\input{05-conclusions.tex}

\subsubsection*{Acknowledgements}

We thank Michael Riis Andersen for his invaluable comments and helpful suggestions.

\bibliography{references}
\bibliographystyle{plainnat}

\clearpage

\end{document}

%% file: 01-introduction.tex
Gaussian processes (GPs) are a family of flexible function distributions defined by a kernel function \citep{rasmussen2006}. The modeling capacity is determined by the chosen kernel. Standard stationary kernels lead to models that underperform in practice. Shallow -- or single layer -- Gaussian processes are often sub-optimal since flexible kernels that would account for non-stationary patterns and long-range interactions in the data are difficult to design and infer \citep{wilson2013,remes2017}. Deep Gaussian processes boost performance by modelling networks of GP nodes \citep{duvenaud2011,sun2018} or by mapping inputs through multiple Gaussian process 'layers' \citep{damianou2013,double-gp}. While more flexible and powerful than shallow GPs, deep Gaussian processes result in degenerate models if the individual GP layers are not invertible, which limits their potential \citep{duvenaud2014}. 

Convolutional neural networks (CNN) are a celebrated approach for image recognition tasks with superior performance \citep{understanding-convnets}. These models encode a hierarchical translation-invariance assumption into the structure of the model by applying convolutions to extract increasingly complex patterns through the layers. 

While neural networks have achieved unparalleled results on many tasks, they have their shortcomings. Effective neural networks require large number of parameters that require careful optimisation to prevent overfitting. %are sensitive to out-of-sample noise. 
Neural networks can often leverage a large number of training data to counteract this problem.
%They require lots of training data to counteract these problems. 
Developing methods that are better regularized and can incorporate prior knowledge would allow us to deploy machine learning methods in domains where massive amounts of data is not available. Conventional neural networks do not provide reliable uncertainty estimates on predictions, which are important in many real world applications. 

The deterministic CNN's have been extended into the probabilistic domain with weight uncertainties \citep{blundell2015}. \citet{gal2016} explored the Bayesian connections of the dropout technique. Neural networks are known to converge to Gaussian processes at the limit of infinite layer width \citep{mackay1992,williams1997, lee2017deep}. \citet{garriga2018} derive a kernel which is equivalent to residual CNNs with a certain prior over the weights. \citet{wilson2016} proposed a hybrid deep kernel learning approach, where a feature-extractor deep neural network is stacked with a Gaussian process predictor layer, learning the neural network weights by variational inference \citep{stochastic-dkl}. 

Recently \citet{conv-gp} proposed the first convolution-based Gaussian process for images with promising performance. They proposed a weighted additive model where Gaussian process responses over image subpatches are aggregated for image classification. The convolutional Gaussian process is unable to model pattern combinations due to its restriction to a single layer. Very recently \citet{deep-conv-kernel} applied convolutional kernels in a deep Gaussian process, however they were unable to significantly improve upon the shallow convolutional GP model. %\citet{tran2018} explore the calibration properties of convolutional GPs and propose a combination of a CNN and a GP to mitigate calibration issues.

In this paper we propose a deep convolutional Gaussian process, which iteratively convolves several GP functions over the image. We learn multimodal probabilistic representations that encode combinations of increasingly complex pattern combinations as a function of depth. Our model is a fully Bayesian kernel method with no neural network component. On the CIFAR-10 dataset, deep convolutions increase the current state-of-the-art GP predictive accuracy from 65\% to 76\%. Our model demonstrates how a purely GP based approach can reach the performance of hybrid neural network GP models.

%% file: 02-background.tex
In this section we provide an overview of the main methods our work relies upon. We consider supervised image classification problems with $N$ examples $\X = \{\x_i\}_{i=1}^N$ each associated with a label $y_i \in \mathbb{Z}$. We assume images $\x \in \R^{H \times W \times C}$ as 3D tensors of size $H \times W \times C$ over $C$ channels, where RGB color images have $C=3$ color channels.

\subsection{Discrete convolutions}

A convolution as used in convolutional neural networks takes a signal, two dimensional in the case of an image, and a tensor valued filter to produce a new signal \citep{deep-learning-book}. The filter is moved across the signal and at each step taking a dot product with the corresponding section in the signal. The resulting signal will have a high value where the signal is similar to the filter, zero where it's orthogonal to the filter and a low value where it's very different from the filter. A convolution of a two dimensional image $\x$ and a convolutional filter $\mathbf{g}$ is defined:
\begin{align}
    (\x \ast \mathbf{g})[i, j] = \sum_{w=0}^{W-1} \sum_{h=0}^{H-1} \x[i + w, j + h] \mathbf{g}[w, h]
\end{align}
$\x[i, j] \in \R^{3}$ and $\mathbf{g}$ is in $\R^{H \times W \times 3}$. Here $H$ and $W$ define the size of the convolutional filter. Typical values could be $H = W = 5$ or $H = W = 3$. Typically multiple convolutional filters are used, each convolved over the input to produce several output signals which are stacked together.

By default the convolution is defined over every location of the image. Sometimes one might use only every other location. This is referred to as the \emph{stride}. A stride of 2 means only every other location $i, j$ is taken in the output. 

\subsection{Primer on Gaussian processes}

Gaussian processes are a family of Bayesian models that characterize distributions of functions \citep{rasmussen-gp}. A zero-mean Gaussian process prior on latent function $f(\x) \in \R$,
\begin{align}
f(\x) &\sim \GP( 0, K(\x,\x'))
\end{align}
defines a \emph{prior} distribution over function values $f(\x)$ with mean and covariance:
\begin{align}
\E[ f(\x)] &= 0 \\
\cov[ f(\x), f(\x')] &= K(\x,\x')
\end{align}
A GP prior defines that for any collection of $n$ inputs $X = (\x_1, \ldots, \x_n)^T$, the corresponding function values \[\f = ( f(\x_1), \ldots, f(\x_n))^T \in \R^n\] follow a multivariate Normal distribution  
\begin{align}
\f \sim \N(\0, \K)
\end{align}
$\K = (K(\x_i, \x_j))_{i,j=1}^n \in \R^{n \times n}$ is the kernel matrix encoding the function covariances. A key property of GPs is that output predictions $f(\x)$ and $f(\x')$ correlate according to the similarity of the inputs $\x$ and $\x'$ as defined by the kernel $K(\x,\x') \in \R$. 

Low-rank Gaussian process functions are constructed by \emph{augmenting} the Gaussian process with a small number $M$ of inducing variables $u_j = f(\z_j)$, $u_j \in \R$ and $\z_j = \R^d$ to obtain the Gaussian function posterior
\begin{align}
    \f | \u,\Z &\sim \N( \underbrace{\K_{\X \Z} \K_{\Z \Z}^{-1} \u}_{\text{predictive mean}}, \underbrace{\K_{\X \X} - \K_{\X \Z} \K_{\Z \Z}^{-1} \K_{\Z \X}}_{\text{predictive covariance}} \, )
%    p(\u) &= \N( \u | \0, \K_{uu}), \label{eq:pu}
\end{align}
where $\K_{\X \X} \in \R^{n \times n}$ is the kernel between observed image pairs $\X$, the kernel $\K_{\X \Z} \in \R^{n \times M}$ is between observed images $\X$ and inducing images $\Z$, and kernel $\K_{\Z \Z} \in \R^{m \times m}$ is between inducing images $\Z$.  \citep{snelson2006sparse}

\subsection{Variational inference}

%The kernel hyperparameters $\theta$ are inferred by maximizing the \emph{evidence},
%\begin{align}
%\log p(\y | \theta) &= \log \mathbb{E}_{p(\f | \theta)} p(\y | \f) \\
% &= \log \N(\y | \0, \K_\theta + \sigma_n^2 I) \\
%  &\propto - \frac{1}{2} \y^T (\K_{nn} + \sigma_y^2 I)^{-1} \y - \frac{1}{2} \log | \K_{nn} + \sigma_y^2 I |, \notag
%  \label{eq:mll}
%\end{align}
%which automatically balances model fit with the square term and the model complexity with the determinant to avoid overfitting (Rasmussen 2006).

Exact inference in a GP entails optimizing the \emph{evidence} $p(\y) = \E_{p(\f)}[p(\y | \f)]$ which has a limiting cubic complexity $O(n^3)$ and is in general intractable. We tackle this restriction by applying stochastic variational inference (SVI) \citep{hensman2015}.

We define a variational approximation
\begin{align}
    q(\u) &= \N( \u | \mathbf{m}, \mathbf{S}) \label{eq:qu} \\
    q(\f) &= \int p(\f | \u) q(\u) d\u \label{eq:qf} \\
     &= \N( \f | \A \mathbf{m}, \K_{ff} - \A (\mathbf{S} - \K_{zz}) \A^T) \notag \\
    \A &= \K_{fz}\K_{zz}^{-1} \notag
\end{align}
with free variational parameters $\m \in \R^m$ and a matrix $\S \succeq 0 \in \R^{m \times m}$ to be optimised. It can be shown that minimizing the Kullback-Leibler divergence $\KL[ q(\u) || p(\u | \y)]$ between the approximative posterior $q(\u)$ and the true posterior $p(\u|\y)$ is equivalent to maximizing the evidence lower bound (ELBO) \citep{vi-for-stats}
\begin{align}
    \mathcal{L} &= \sum_{i=1}^n \mathbb{E}_{q(f_i)} [ \log p(y_i | f_i) ] - \KL[ q(\u) || p(\u)] \label{eq:elbo}
\end{align}
The variational expected likelihood in $\mathcal{L}$ can be computed using numerical quadrature approaches \citep{scalable-gp}.
%for Probit classification \citep{hegde2018}, 

%The variational bound \eqref{eq:elbo} along with equations \cref{eq:qu,eq:qf} require the ability to compute the image self-kernels $K_{ff}(\x,\x)$, the kernel between images and inducing points $K_{fz}(\x, \z)$, and the kernel between inducing points $K_{zz}(\z, \z')$.

%% file: 03-methods.tex
\begin{figure*}[ht]
    \centering
    \includegraphics[width=0.9\linewidth]{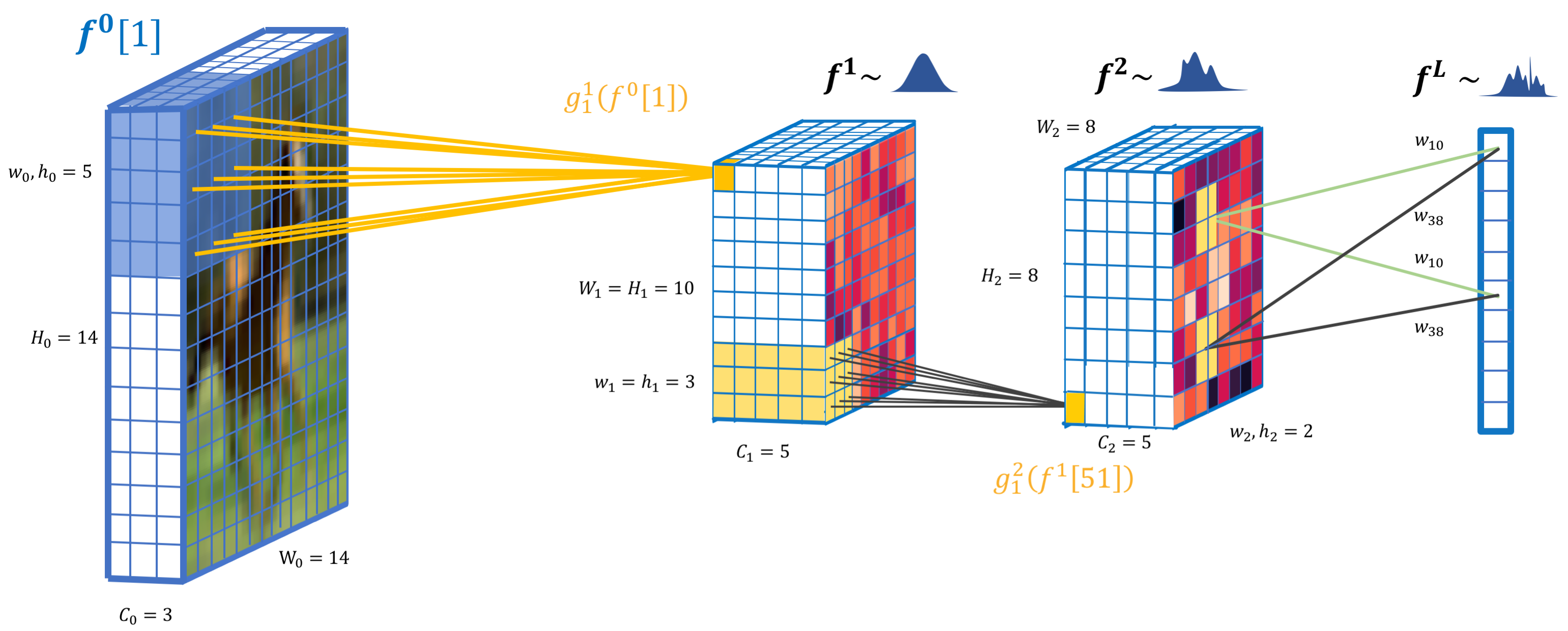}
    \caption{A three layer deep convolutional gaussian process. First we construct an intermediate probabilistic representation of size $W_1 \times H_1 \times C_1$. We map this probabilistic representation through another convolutional GP layer yielding a representation of size $W_2 \times H_2 \times C_2$. Finally, we classify using a GP with a convolutional kernel by summing over patches of the intermediate representation.}
    \label{fig:overview}
\end{figure*}
\section{Deep convolutional Gaussian process}
\label{sec:methods}

In this section we introduce the deep convolution Gaussian process. We stack multiple convolutional GP layers followed by a GP classifier with a convolutional kernel.
\subsection{Convolutional GP layers}

We assume an image representation $\f^\l_c \in \R^{H_\l \times W_\l}$ of width $W_\l$ and height $H_\l$ pixels at layer $\l$. We collect $C_\l$ \emph{channels} into a 3D tensor $\f^\l = (\f^\l_1, \ldots, \f^\l_C) \in \R^{H_\l \times W_\l \times C_\l}$, where the channels are along the depth axis. The input image $\f^0 = \x$ is the $W_0 \times H_0 \times C_0$ sized representation of the original image with C color channels. For instance MNIST images are of size $W = H = 28$ pixels and have a single $C=1$ grayscale channel.

We decompose the 3D tensor $\f^\l$ into \emph{patches} $\f^\l[p] \in \R^{w_\l \times h_\l \times C_\l}$ containing all depth channel. $h_\l$ and $w_\l$ are the height and width of the image patch at layer $\l$. We index patches by $p \in \mathbb{Z} < H_\l W_\l$. $H_\l$ and $W_\l$ denotes the height and width of the output of layer $\l$. We compose a sequence of layers $\f^\l$ that map the input image $\x_i$ to the label $\y_i$:
\begin{align}
    \underbrace{\x_i = \f^0}_{W_0 \times H_0 \times 3} \xrightarrow{g^1} \underbrace{\f^1}_{W_1 \times H_1 \times C_1} \cdots \xrightarrow{g^L} \underbrace{\f^L}_{C_y} \approx \underbrace{\y_i}_{\{0,1\}^{C_y}}
\end{align}
Layers $\f^\l$ with $\l \ge 1$ are random variables with probability densities $p(\f^\l)$.

%Let $\f^{l-1}[1, 1], \f^{l-1}[1, 2], ..., \f^{l-1}[P_H, P_W]$ be patches of the input. Each patch has size $h \times w \times c$ where c corresponds to the color channel. At the first layer $\f^0 = \x$ is the original input image. $\{\Z^\l\}_{\l=1}^L$ are inducing points each of the same size as our image patches. Our model consists of several layers, each having their own inducing points. $\Z^\l$ denotes the inducing points at layer $l$. $\f^\l$ represents the output of layer $l$.

We construct the layers by applying \emph{convolutions} of \emph{patch response} functions $\g_c^\l: \R^{w_{\l-1}\times h_{\l-1} \times C_{\l-1}} \rightarrow \R$ over the input one patch at a time producing the next layer representation:
\begin{align}
\f^{\l}[p] =
\begin{bmatrix}
g_1^{\l}( \f^{\l-1}[p]) \\ \vdots \\ g_C^{\l}( \f^{\l-1}[p])
\end{bmatrix} \in \R^C
\end{align}
Each individual patch response $g^\l(\f^{\l-1}[p])$ is a $1 \times 1 \times C$ pixel stack. By repeating the patch responses over the $P_{\l-1} = H_\l \times W_\l$ patches we form a new $W_\l \times H_\l \times C_\l$ representation $\f^{\l} = ( \f^{\l}[1], \ldots, \f^{\l}[P_{\l-1}])$ (See Figure \ref{fig:overview}). 

We model the $C$ patch responses at each of the first $L-1$ layers as independent GPs with shared prior
\begin{align}
g_c^{\l}( \f^{\l-1}[p] ) &\sim \GP\big( 0, k(\f^{\l-1}[p],\f'^{\l-1}[p'] ) \big)
\end{align}
for $c = 1,\ldots, C$. The kernel $k(\cdot,\cdot)$ measures the similarity of two image patches. The standard property of Gaussian processes implies that the functions $g_c^\l$ output similar responses for similar patches. 

\begin{figure*}[t]
    \centering
    \includegraphics[width=\linewidth]{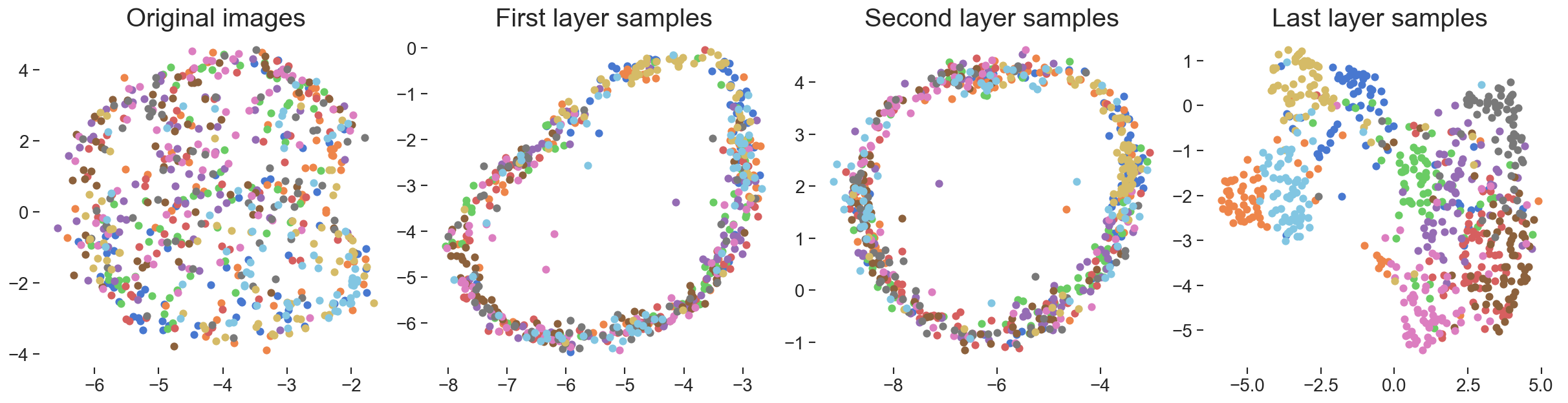}
    \caption{UMAP embeddings \citep{umap} of the CIFAR-10 images and representations after each layer of the deep convolutional GP model. The colors correspond to different classes in the classification problem.}
    \label{fig:transform}
\end{figure*}

\begin{figure*}[t]
    \centering
    \includegraphics[width=0.94\linewidth]{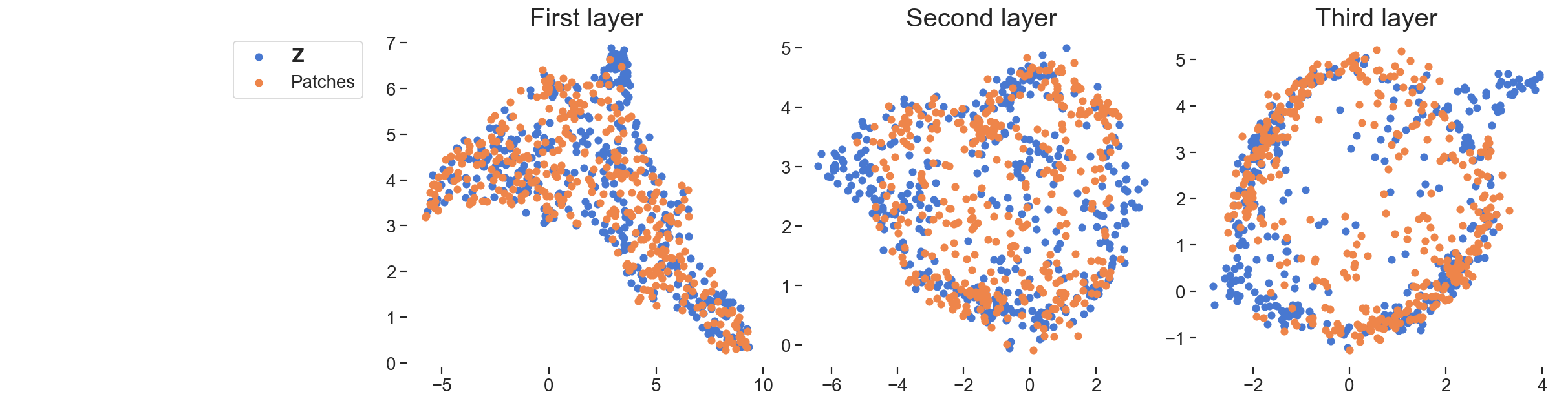}
    \caption{UMAP embeddings of randomly selected patches of the input to the layer and learned inducing points of the fitted three layer model on CIFAR-10.}
    \label{fig:inducing}
\end{figure*}

For example, on MNIST where images have size $28 \times 28 \times 1$ using patches of size $5 \times 5 \times 1$, a stride of 1 and $C=10$ patch response functions, we obtain a representation of size $24 \times 24 \times 10$ after the first layer (height and width $W_1 = H_1 = (28 - 5) / 1 + 1$). This is passed on to the next layer which produces an output of size $20 \times 20 \times 10$.

We follow the sparse GP approach of \cite{hensman2015} and augment each patch response function by a set of $M$ inducing patches $\z^\l$ in the patch space $\R^{h_{\l-1} \times w_{\l-1} \times C_{\l-1}}$ with corresponding responses $u_c^\l$. Each layer contains $M_\l$ inducing patches $\Z^\l = (\z_1^\l, \ldots, \z_M^\l)$ which are shared among the $C$ patch response functions within that layer. Each patch response function has separate inducing responses $\u_c^\l = ( u_{c1}^\l, \ldots, u_{cM}^\l)$ which associate outputs to each inducing patch. We collect these into a matrix $\U^\l$. 

%The $M$ filters $\Z^\l = (\z_1^\l, \ldots, \z_M^\l)$ at each layer are shared amongst patch response functions within a layer. Each patch response function will have separate $\u_c^\l = ( u_{c1}^\l, \ldots, u_{cM}^\l)$ collected into the matrix $\U^\l$.

The conditional patch responses are
\begin{align}
g_c^{\l} | \f^{\l-1}, \u^\l_c, \Z^\l  &\sim \N(\bmu,\Sigma) \label{eq:indconv}\\ 
\bmu &= \K_{\f^{\l-1} \Z^\l} \K_{\Z^\l \Z^\l}^{-1} \u^\l_c \notag \\
\Sigma &= \K_{\f^{\l-1} \f^{\l-1}} - \K_{\f^{\l-1} \Z^\l} \K_{\Z^\l \Z^\l}^{-1} \K_{\Z^\l \f^{\l-1}},\notag
\end{align}
where the covariance between the input and the inducing variables are 
\begin{align*}
    K(\f^{\l-1}, \Z^\l) &= \begin{bmatrix}
        k(\f^{\l-1}[1], \z_1^\l) & \cdots & k(\f^{\l-1}[1], \z_M^\l) \\
        \vdots & \ddots & \vdots \\
        k(\f^{\l-1}[P], \z_1^\l) & \cdots & k(\f^{\l-1}[P], \z_M^\l)
    \end{bmatrix}
\end{align*}
a matrix of size $P_\l \times M_\l$ that measures the similarity of all patches against all filters $\z^\l$. We set the base kernel $k$ to be the RBF kernel. For each of the $C$ patch response functions we obtain one output image channel. 

In contrast to neural networks, the Gaussian process convolutions induce probabilistic layer representations. The first layer $p(\f^1 | \f^0, \U^1, \Z^1)$ is a Gaussian directly from \eqref{eq:indconv}, while the following layers follow non-Gaussian distributions $p(\f^{\l+1} | \U^{\l+1}, \Z^{\l+1})$ since we map all realisations of the random input $\f^{\l}$ into Gaussian outputs $\f^{\l+1}$.

%We convolve a gaussian process over the input signal one patch at a time to produce a new representation of the image. Formally, we define $J$ \emph{patch response} functions $g_j^l: \R^{h \times w \times c} \mapsto \R$ per layer $l$ upon which we set GP priors 
%\begin{align}
%    g^l_j[ &= g^{lk}(\f^{l-1}[i, j]) \sim GP(\mathbf{0}, K(\f^{l-1}[i, j], \f'^{l-1}[i,j]))
%\end{align}

%$g_{ij}^{lk}$ is the $k$:th patch response to the image patch $\x[i, j]$ at layer $l$. Each patch response function at each layer is associated with a variational distribution $q_k(\u^{l}) = \N(\m^{l}_k, \S^{l}_k)$. 

%The stride defines how far apart patches are from each other. With a stride of one we move the receptive field of the patch response function by one pixel at each step. 

%Following the sparse GP approach we get the conditional distribution 
%\begin{align}
%    p(\f^{lk} | \m^l_k, \mathbf{S^l_k}) &= \N(\A\m^l_k, \A(\K_{zz} - \mathbf{S}^l_k)\A^T) \\
%    \A &= \K_{\f^{l-1}\z^l}\K_{\z^l\z^l}^{-1}
%\end{align}
%
%where

%For each of the $K$ patch response functions we obtain one output image each placed in the `color channel` of the layer output. The next layer extracts patches cutting through all channels. 

\begin{figure*}[t]%
    \centering
    \begin{subfigure}[c]{.33\textwidth}
        \centering
        \includegraphics[width=\linewidth]{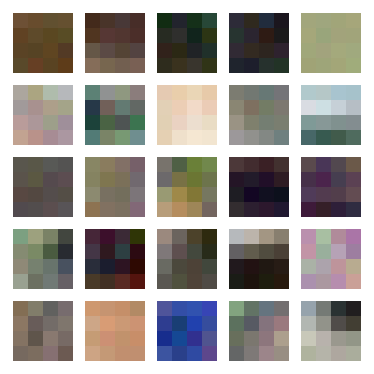}
        \caption{Layer 1}
        \label{fig:Z1}
    \end{subfigure}%
    \hfill
    \begin{subfigure}[c]{.33\textwidth}
        \centering
        \includegraphics[width=\linewidth]{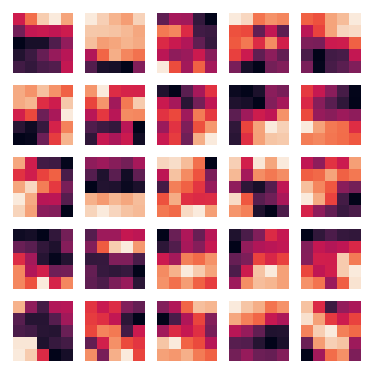}
        \caption{Layer 2}
        \label{fig:Z2}
    \end{subfigure}%
    \begin{subfigure}[c]{.33\textwidth}
        \centering 
        \includegraphics[width=\linewidth]{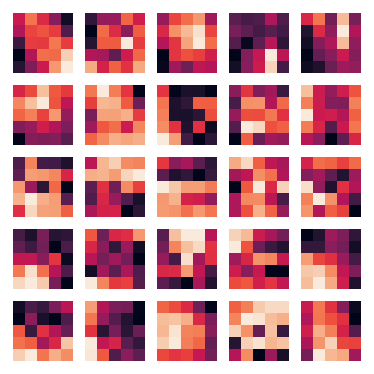}
        \caption{Layer 3}
        \label{fig:Z3}
    \end{subfigure}
    \caption{Example inducing points $\Z$ pictured from all three layers from the CIFAR-10 experiment. The first layer inducing points channels correspond to color channels and are thus in color. For layers 2 and 3 only a single channel is visualized.}
    \label{fig:Z}
\end{figure*}

\subsection{Final classification layer}
\label{sec:background:conv-gp}

As the last layer of our model we aggregate the output of the convolutional layers using a GP with a weighted convolutional kernel as presented by \citet{conv-gp}. We set a GP prior on the last layer patch response function 
\begin{align}
g^L\big(\f^{L-1}[p]\big) &\sim \GP(0, K( \f^{L-1}[p], {\f'^{L-1}}[p'] ) ).
\end{align}
with weights for each patch response. We get an additive GP
\begin{align*}
    \f^L &= g^L(\f^{L-1}) = \sum_{p=1}^P w_p g^L(\f^{L-1}[p]) \\
    &\sim \GP\Bigg(0, \underbrace{\sum_{p=1}^P\sum_{p'=1}^P w_p w_{p'} k(\f^{L-1}[p], \f'^{L-1}[p'])}_{K(\x,\x')}\Bigg),
\end{align*}
where the kernel $K(\f^{L-1}, \f'^{L-1}) = \w^T \K \w$ is the weighted average patch similarity of the final tensor representation $\f^{L-1}$. $\w \in \R^P$. The matrix $\K$ collects all patch similarities $K( \f^{L-1}[p], \f'^{L-1}[p'])$. The last layer has one response GP per output class $c$.

As with the convolutional layers the inducing points live in the patch space of instead of in the image space. The inter-domain kernel is 
\begin{align}
K(\f^{L-1},\z^{L}) &= \sum_{p=1}^P w_p K(\x[p],\z^{L}) \\
 &= \w^T \bk(\f^{L-1},\z^{L}).
\end{align}
The kernel $\bk(\f^{L-1},\z^{L}) \in \R^{P}$ collects all patch similarities of a single image $\f^{L-1}$ compared against inducing points $\z^{L}$. The covariance between inducing points is simply $K(\z^{L}, \z'^{L})$. We have now defined all kernels necessary to evaluate and optimize the variational bound \eqref{eq:elbo}.

%The Figure \ref{fig:transform} visualises the embedding

\subsection{Doubly stochastic variational inference}

\begin{table*}[t]
\centering
\medskip
%\begin{tabularx}{\linewidth}{*{3}{p{.33\linewidth}}}
\begin{tabular}{l cc cc r}
\toprule
& & Inducing & \multicolumn{2}{c}{Test accuracy} & \\
%\cmidrule(r{31}){2-3}
\cmidrule{4-5}
\textbf{Gaussian process models} & Layers & points & \textbf{MNIST} & \textbf{CIFAR-10} & \textbf{Reference} \\ 
\midrule
RBF AutoGP & $1$ & $200$ & $98.29^{(*)}$ & $55.05^{(*)}$ & \citet{krauthautogp} \\
Multi-channel conv GP & $1$ & $1000$ & $98.83^{(*)}$ & $64.6^{(*)}$ & \citet{conv-gp} \\
%\cmidrule{1-4}
% DeepCGP & $1$ & $1000$ & 98.58 & 59.94  & current work \\
DeepCGP & $1$ & $384$ & 98.38 & 58.65  & current work \\
DeepCGP & $2$ & $2 \times 384$ & 99.24 & 73.85 & " $\qquad$ \\ 
DeepCGP & $3$ & $3 \times 384$ & \textbf{99.44} & \textbf{75.89} & " $\qquad$ \\ 
%\bottomrule
\midrule
\textbf{Neural network models} & Layers & \# params & & & \\
\midrule
Deep kernel learning & 5 & 2.3M .. 4.6M & $99.2^{(*)}$ & $77.0^{(*)}$ & \citet{stochastic-dkl} \\ 
DenseNet & 250 & 15.3M &  N/A & $94.81^{(*)}$ & \citet{densenet} \\
\bottomrule
\end{tabular}
\caption{Performance on MNIST and CIFAR-10. Our method, the deep convolutional Gaussian process, is denoted DeepCGP. Asterisk $^{(*)}$ indicates results taken from the respective publications, which are directly comparable due to standard data folds. Other results are run using our implementation. The neural network based results are listed for completeness.  %\textsuperscript{0} run by us using our implementation. %\textsuperscript{1} taken from \citep{conv-gp}. \textsuperscript{2} taken from \citep{stochastic-dkl}. \textsuperscript{3} taken from \citep{densenet}.
}
\label{table:perf}

\end{table*}

The deep convolutional Gaussian process is an instance of a deep Gaussian process with the convolutional kernels and patch filter inducing points. We follow the doubly stochastic variational inference approach of \citet{double-gp} for model learning. The key idea of doubly stochastic inference is to draw samples from the Gaussian
\begin{align}
    \tilde{\f}^{\l}_i \sim p(\f_i^{\l} | \tilde{\f}_i^{\l-1}, \U^{\l},\Z^{\l})
\end{align}
through the deep system for a single input image $\x_i$.

The inducing points of each layer are independent. We assume a factorised likelihood
\begin{align}
    p( \mathbf{Y} |  \mathbf{F}^L ) &= \prod_{i=1}^N p(\y_i | \f^L_i ) 
\end{align}
and a true joint density
\begin{align}
    p( \{ \f^\l,\U^\l \}_\l ) &= \prod_{\l=1}^{L} p(\f^{\l} | \f^{\l-1}, \U^{\l}, \Z^{\l} ) p(\U^\l) \\
    p(\U^\l) &= \prod_{c=1}^C \N(\u_c^\l | \0, \K_{\Z^\l \Z^\l}).
\end{align}
The evidence framework \cite{mackay1992} considers optimizing the evidence,
\begin{align}
    p(\mathbf{Y}) &= \E_{p(\F)} p(\mathbf{Y} | \mathbf{F} ).
\end{align}
Following the variational approach we assume a variational joint model
\begin{align}
    q(\U^\l) &= \prod_{c=1}^C \N(\u_c^\l | \m_c^\l, \S_c^\l) \\
    q\big( \{ \f^\l, \U^\l \}_{\l} \big) &= \prod_{\l=1}^{L} p(\f^{\l} | \f^{\l-1}, \U^\l, \Z^\l ) q(\U^\l).
\end{align}
The distribution of the layer predictions $\f^{\l}$ depends on current layer inducing points $\U^\l,\Z^\l$ and representation $\f^{\l-1}$ at the previous layer. By marginalising the variational approximation $q(\U^\l)$ we arrive at the factorized variational posterior of the last layer for individual data point $\x_i$,
\begin{align}
    q(\f_i^L ; \{ \m^\l, \S^\l, \Z^\l \}_\l ) &= \prod_{\l=1}^{L-1} \int q(\f_i^{\l} | \f_i^{\l-1}, \m^l, \S^\l, \Z^\l ) d\f_i^{\l},
\end{align}
where we integrate all paths $(\f_i^1, \ldots, \f_i^L)$ through the layers defined by the filters $\Z^\l$, and the parameters $\m^\l,\S^\l$. Finally, the doubly stochastic evidence lower bound (ELBO) is
\begin{align}
\log p(\mathbf{Y}) &\ge \sum_{i=1}^N \E_{ q(\f_i^L ; \{ \m^\l, \S^\l, \Z^\l \}_\l ) } [ \log p(\y_i | \f_i^L) ] \\
 & \quad - \sum_{\l=1}^L \KL[ q(\U^\l) || p(\U^\l)]. \notag
\end{align}
The variational expected likelihood is computed using a Monte Carlo approximation yielding the first source of stochasticity. The whole lower bound is optimized using stochastic gradient descent yielding the second source of stochasticity.

The Figure \ref{fig:transform} visualises representations of CIFAR-10 images over the deep convolutional GP model. Figure \ref{fig:inducing} visualises the patch and filter spaces of the three layers, indicating high overlap. Finally, Figure \ref{fig:Z} shows example filters $\z$ learned on the CIFAR-10 dataset, which extract image features.

\paragraph{Optimization} All parameters $\{\m_\l\}_{\l=1}^{L}$, $\{\mathbf{S}_\l\}_{\l=1}^{L}$, $\{\Z^l\}_{\l=1}^{L}$, the base kernel RBF lengthscales and variances and the patch weights for the last layer are learned using stochastic gradient Adam optimizer \citep{adam} by maximizing the likelihood lower bound. We use one shared base kernel for each layer.

%In the proposed infinitely deep differential Gaussian processes the choice of divergence-free kernel guarantees rank-preserving, non-degenerate warpings. However, our experiments show that standard rank-reducing Gaussian kernel outperforms the div-free kernel, even though it leads to degeneracy. Hence, the infinitely deep Gaussian process seems to, remarkably, benefit instead of suffer from rank-reducing covariances. We argue that reducing the rank of the model over time is compatible with the goal of finding latent patterns from complex data.

%\paragraph{Tensor representation.}
%We do not compress the layers

%\paragraph{Stochastic inference.}
%Why inference is difficult, where are the bottleneceks. We could try normalizing flows, Stein, SG-HMC as future work

%% file: 04-experiments.tex
% DKL has fully connected DNN's of  (D,1000,1000,500,50,C) nodes, where "D" is the input dimensions, and "C" is output classes. 
% with CIFAR they don't use CNN's but this simple DNN where D=32*32*3=3072 and C=10, with MNIST the D=28*28=784
% With CIFAR they have 3072*1000 + 1000*1000 + 1000*500 + 500*50 + 50*10 = 4597500 weights
% With MNIST they have 784*1000 + 1000*1000 + 1000*500 + 500*50 + 50*10 = 2309500 weights

\begin{figure*}[t]%
\centering
    \begin{subfigure}[c]{.48\textwidth}
        \centering
        \includegraphics[width=\linewidth]{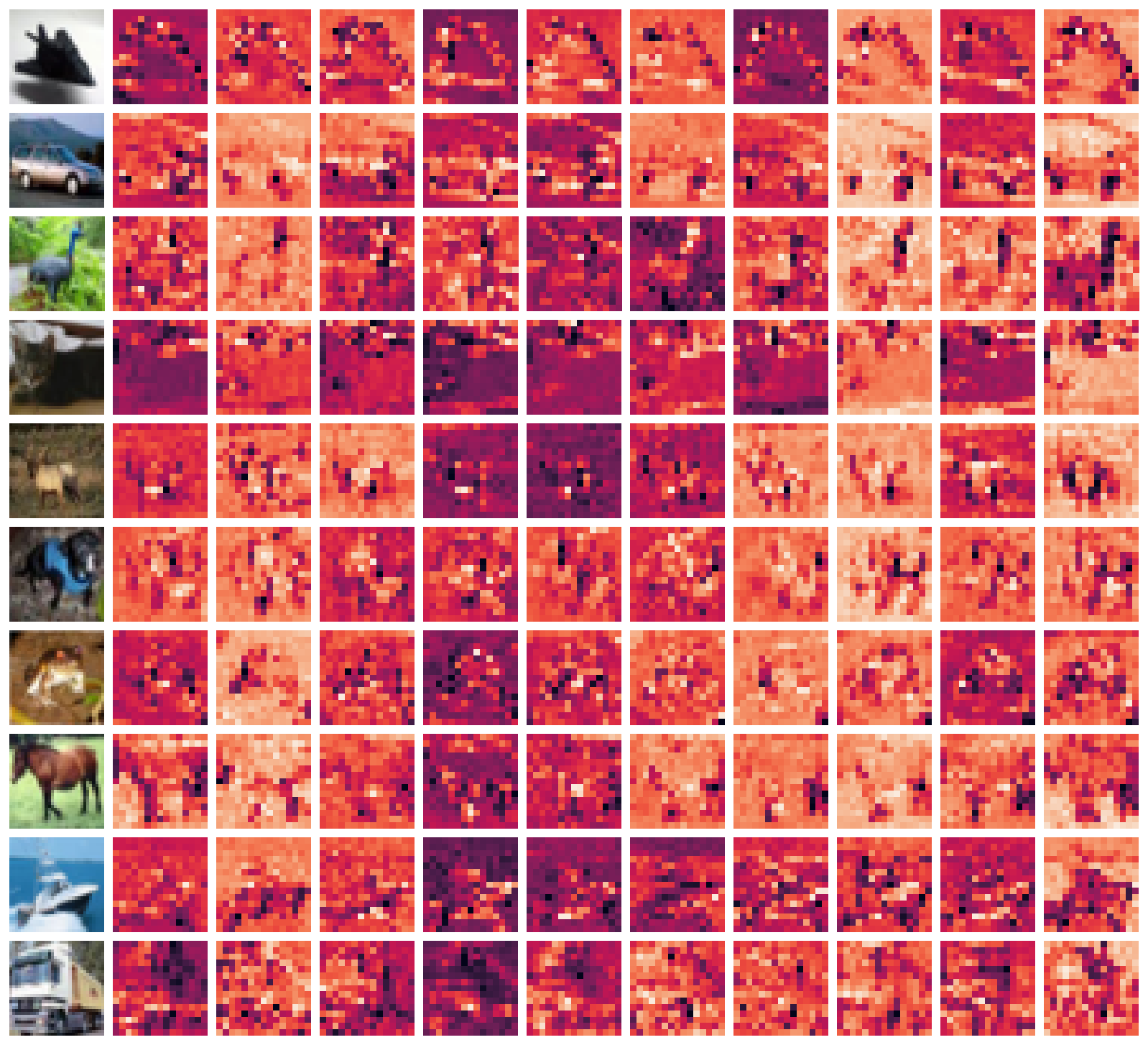}
        \caption{Samples from the first layer.}
        \label{fig:activations1}
    \end{subfigure}%
    \hfill
    \begin{subfigure}[c]{.48\textwidth}
        \centering
        \includegraphics[width=\linewidth]{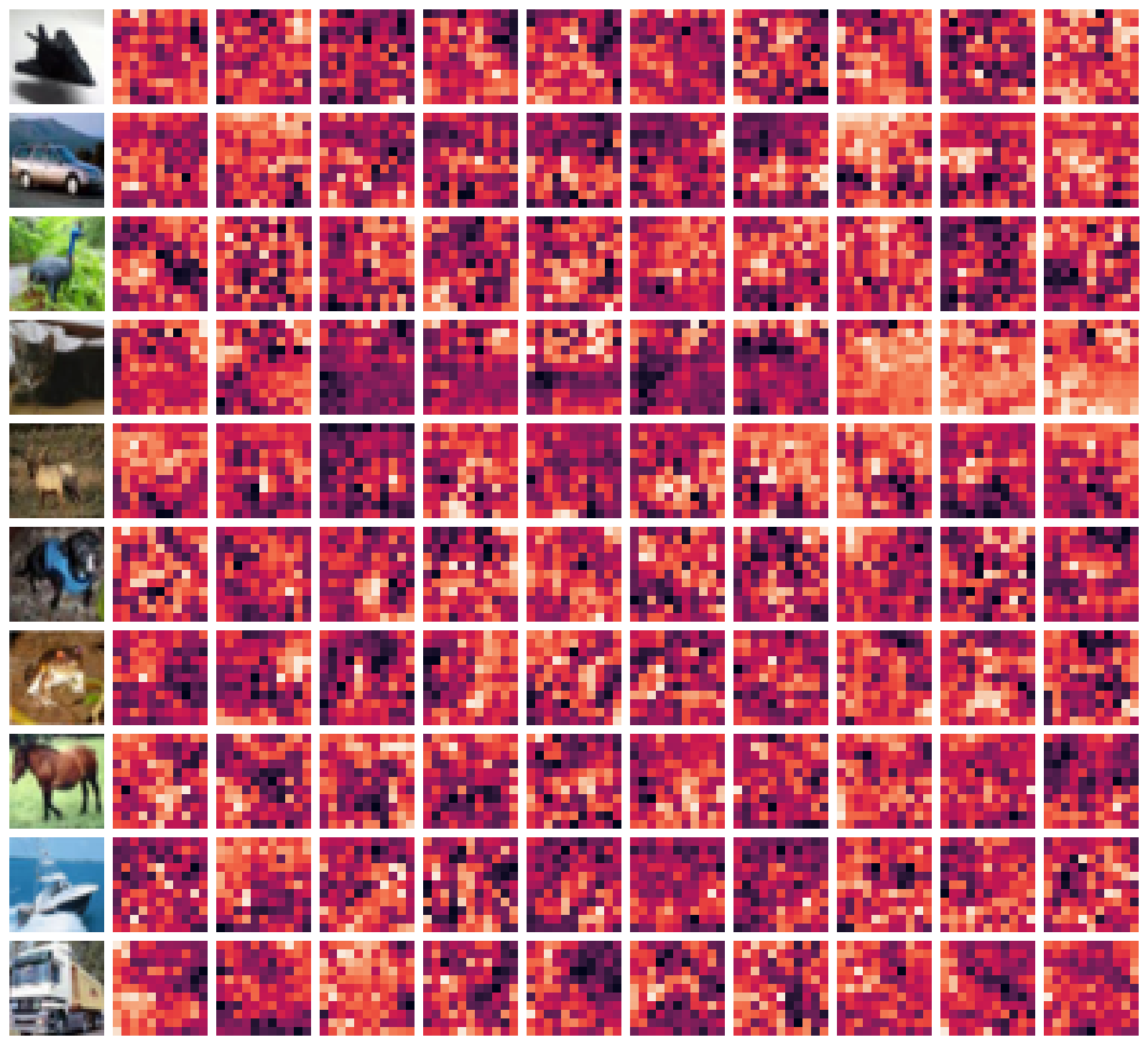}
        \caption{Samples from the second layer.}
        \label{fig:activations2}
    \end{subfigure}%
    \caption{(a) and (b) show samples the first two layers of the three layer model. Rows corresponds to different test inputs and columns correspond to different patch response functions, which are realisations of the layer GPs. The first column shows the input image. The first layer seems to learn to detect edges, while the second layer appears to learn more abstract correlations of features and the representation produced no longer resembles the input image, indicating high-level feature extraction.}
    \label{fig:activations}
\end{figure*}

We compare our approach on the standard image classification benchmarks of MNIST and CIFAR-10 \citep{cifar-10}, which have standard training and test folds to facilitate direct performance comparisons. MNIST contains 60,000 training examples of $28 \times 28$ sized grayscale images of 10 hand-drawn digits, with a separate 10,000 validation set. CIFAR-10 contains 50,000 training examples of RGB colour images of size $32 \times 32$ from 10 classes, with 5,000 images per class. The images represents objects such as airplanes, cats or horses. There is a separate validation set of 10,000 images. We preprocess the images for zero mean and unit variance along the color channel.

\begin{figure}[th]
    \centering
    \includegraphics[width=\columnwidth]{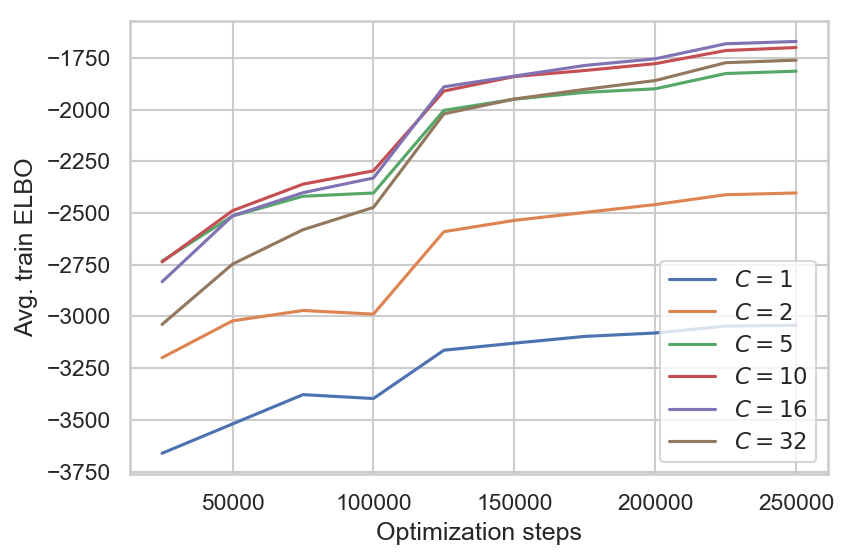}
    \caption{Expected evidence lower bound computed on the training set using a two layer model for different amounts of patch response functions. The models with 10 and 16 patch response functions seem to perform the best. Models with one or two patch response functions struggle to explain the data even though they have the same amount of inducing points.}
    \label{fig:train_log_likelihood}
\end{figure}

We compare our model primarily against the original shallow convolutional Gaussian process \citep{conv-gp}, which is currently the only convolutional Gaussian process based image classifier. We also consider the performance of the hybrid neural network GP approach of \citet{stochastic-dkl}. For completeness we report the performance of a state-of-the-art CNN method DenseNet \citep{densenet}.
%however challenging CNN's is out of scope for this paper.

%against other approaches. Namely: deep kernel learning \citep{stochastic-dkl}, a convolutional neural network approach \citep{densenet} and the convolutional gaussian process \citep{conv-gp}.

%In all experiments we preprocess the images to have zero mean and unit standard deviation along each color channel. 

\paragraph{Implementation.}
Our TensorFlow \citep{abadi2016tensorflow} implementation is compatible with the GPflow framework \newline \citep{gpflow} and freely available \textbf{\href{https://github.com/kekeblom/DeepCGP}{online}} \footnote{\href{https://github.com/kekeblom/DeepCGP}{https://github.com/kekeblom/DeepCGP}}. We leverage GPU accelerated computation, 64bit floating point precision, and employ a minibatch size of 32.
We start the Adam learning rate at $0.01$ and multiply it by $0.1$ every 100,000 optimization steps until the learning rate reaches 1e-5. We use $M=384$ inducing points at each layer. We set a stride of $2$ for the first layer and $1$ for all other layers. The convolutional filter size is 5x5 on all layers except for the first layer on CIFAR-10 where it is 4x4. This is to make use of all the image pixels using a stride of 2.

\paragraph{Parameter initialization.}
Inducing points $\Z$ are initialized by running $k$-means with $M$ clusters on image patches from the training set. The variational means $\m$ are initialised to zero. $\S$ are initialised to a tiny variance kernel prior $10^{-5} \cdot K_{\Z \Z}$ following \citet{double-gp}, except for the last layer where we use $K_{\Z \Z}$. For models deeper than two layers, we employ iterative optimisation where the first $L-2$ layers and layer $L$ are initialised to the learned values of an $L-1$ model, while the one additional layer added before the classification layer is initialised to default values.

\subsection{MNIST and CIFAR-10 results}

Table \ref{table:perf} shows the classification accuracy on MNIST and CIFAR-10. Adding a convolutional layer to the weighted convolutional kernel GP improves performance on CIFAR-10 from 58.65\% to 73.85\%. Adding another convolutional layer further improves the accuracy to 75.9\%. On MNIST the performance increases from $1.42\%$ error to $0.56\%$ error with the three-layer deep convolutional GP. 

% The performance of a state-of-the-art CNN method DenseNet is superior, while the model is also much more complex with hundreds of layers and millions of parameters. 
The deep kernel learning method uses a fully connected five-layer DNN instead of a CNN, and performs similarly to our model, but with much more parameters. 

%\paragraph{Model structure.}
Figure \ref{fig:activations} shows a single sample for 10 image class examples (rows) over the 10 patch response channels (columns) for the first layer (panel a) and second layer (panel b). The first layer indicates various edge detectors, while the second layer samples show the complexity of pattern extraction. The row object classes map to different kinds of representations, as expected.

Figure \ref{fig:transform} shows UMAP embedding \cite{umap} visualisations of the image space of CIFAR-10 along with the structure of the layer representations $\f_i^\l$ for three layers. The original images do not naturally cluster into the 10 classes (a). The DCGP model projects the images to circle shape with some class coherence in the intermediate layers, while the last layer shows the classification boundaries. An accompanying Figure \ref{fig:Z} shows the learned inducing filters and layer patches on CIFAR-10. Some regions of the patch space are not covered by filters, indicating uninformative representations.

Figure \ref{fig:train_log_likelihood} shows the effect of different channel numbers on a two layer model. The ELBO increases up to $C=16$ response channels, while starts to decrease with $C=32$ channels. A model with approximately $C=10$ channels indicates best performance. 

%shows average training example log likelihood for models with a different amount of patch response functions at the convolutional layers. Showing that the models with 10 and 16 patch response functions perform the best.

%% file: 05-conclusions.tex
We presented a new type of deep Gaussian process with convolutional structure. The convolutional GP layers gradually linearize the data using multiple filters with nonlinear kernel functions. Our model greatly improves test results on the compared classification benchmarks compared to other GP-based approaches, and approaches the performance of hybrid neural-GP methods. The performance of our model seems to improve as more layers are added. 
%While we did not yet achieve the level of current state-of-the-art deep neural networks, we are certainly getting closer.

We did not experiment with using a stride of 1 at the first layer. Neither did we try models with 4 or more layers. The added complexity comes with an increased computational cost and we were thus limited from experimenting with these improvements. We believe that both of these enhancements would increase performance.

Deep Gaussian process models lead to degenerate covariances, where each layer in the composition reduces the rank or degrees of freedom of the system \citep{duvenaud2014}. In practise the rank reduces via successive layers mapping inputs to identical values, effectively merging inputs and resulting in rank-reducing covariance matrix with repeated rows and columns. To counter this pathology \citet{double-gp} proposed rank-preserving deep model by pseudo-monotonic layer mappings with GP priors $f(\x) \sim \GP(\x, k)$ with identity means $\E[f(\x)] = \x$. In contrast we employ zero-mean patch response functions. Remarkably we do not experience rank degeneracy, possibly due to the multiple channel mappings and the convolution structure.

%We do not compress the layers
%Why inference is difficult, where are the bottleneceks. We could try normalizing flows, Stein, SG-HMC as future work

There are several avenues for improved efficiency and modelling capacity. The Stochastic Gradient Hamiltonian Monte Carlo approach \citep{ma2015complete} has proven efficient in deep GPs \citep{sghmc-gp} and in GANs \citep{saatci2017bayesian}. Another avenue for improvement lies in kernel interpolation techniques \citep{wilson2015kernel,evans2018scalable} which would make inference and prediction faster. We leave these directions for future work.

%Results could be improved should more efficient methods of inference in deep GPs be made available. Future research could try alternative deep GP inference approaches. The Stochastic Gradient Hamiltonian Monte Carlo approach taken by \cite{sghmc-gp} might be worth considering.

%One could also imagine exploring the use of faster GP approximations such as the kernel interpolation approach presented by \cite{gardner2018product} to alleviate the computational burden imposed by our approach.